\documentclass[11pt]{article}

\usepackage[final]{acl}

\usepackage{times}
\usepackage{latexsym}

\usepackage[T1]{fontenc}
\usepackage[utf8]{inputenc}

\usepackage{microtype}
\usepackage{hyperref}
\usepackage{booktabs}  % for \toprule, \midrule, \bottomrule
\usepackage{float} 
\usepackage{csquotes}
\usepackage{amsmath}
\usepackage{amssymb}
\usepackage{inconsolata}

\usepackage{graphicx}

\usepackage{linguex}
\usepackage[dvipsnames]{xcolor}

\title{SAGE (Subjectivity-Aware Generation Estimator):
A Linguistic Framework for AI-Generated Review Detection
\thanks{Preprint version, October 2026.
This manuscript is currently under review.}}

\author{
  \begin{tabular}{c}
    \textbf{Gonzalo Freijedo Aduna$^{1}$,
    Charles Yaiche$^{2}$,
    Patrick Bas$^{2}$,
    Alda Mari$^{1}$} \\[6pt]
    $^{1}$Institut Jean Nicod - CNRS ENS/EHESS PSL University, Paris, France \\
    $^{2}$ UMR 9189 CRIStAL, Univ. Lille, CNRS, Centrale Lille\\
    \texttt{gonzalo.freijedo.aduna@ens.psl.eu} \\
    \texttt{charles.yaiche.lf@gmail.com}\\
    \texttt{patrick.bas@cnrs.fr} \\
    \texttt{alda.mari@cnrs.fr}
  \end{tabular}
}

\begin{document}
\maketitle
\begin{abstract}

We propose a simple, linguistically motivated approach to detecting AI-generated peer reviews based on the inherently human trait of subjectivity. We capture subjectivity at two levels: the semantic level of evaluative language and modality and the pragmatic level of speech acts. Using features derived from speech act annotations and model-predicted subjectivity scores, we train a lightweight classifier that achieves an AUROC of 0.9566 with only these linguistic features. Combining its output with the state-of-the-art detector \textit{Sem-Detect} raises the TPR at 1\% FPR of \textit{Sem-Detect} on the APRDB benchmark from 0.874 to 0.985, outperforming the current state of the art. Our results show that linguistic features of subjectivity provide an explainable source of information for detecting AI-generated peer reviews and can be readily integrated into existing detection methods. The code and models are publicly available on GitHub (\url{https://github.com/Arys02/generated_text_detection_hub}) and Hugging Face (\url{https://huggingface.co/Arys02}),
respectively.

\end{abstract}

\section{Introduction} \label{Intro}

AI-generated text detection is becoming one of the most challenging problems in natural language processing, as increasingly capable language models produce text that closely resembles human writing and can evade existing detection methods \citep{fraserDetectingAIGeneratedText2025, wuSurveyLLMGeneratedText2025a}. This challenge is particularly relevant in scientific research, where the increasing adoption of LLMs is reshaping both the production and evaluation of scientific works. \citet{kusumegiScientificProductionEra2025} show that adopting LLMs to draft manuscripts is associated with a significant increase in paper production. The resulting growth in submissions also increases the burden on peer reviewers, potentially encouraging, in turn, the use of LLMs for reviewing. This makes detection especially pressing for peer reviews, which inform decisions on research evaluation, promotion, and selection, and where falsely identifying human-written text as AI-generated can have significant consequences for researchers' reputations and careers.

\begin{figure*}[th]
  \centering
  \includegraphics[width=\textwidth]{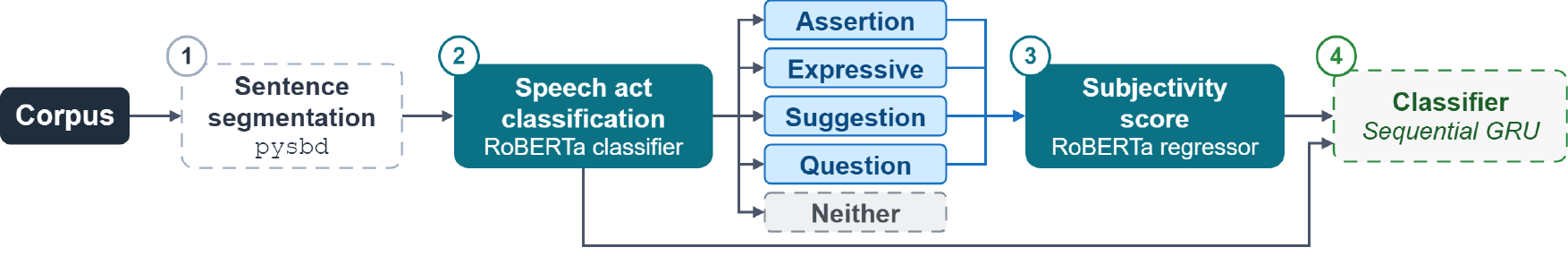}
  \caption{Overview of the {\it SAGE} pipeline. Speech act classification and subjectivity estimation are performed by two independently fine-tuned RoBERTa models: a five-class classifier and a regressor producing continuous scores on a 1--5 scale (Section~\ref{ann-pipe}).}
  \label{pipeline}
\end{figure*}

Several recent studies have recently addressed the detection of AI-generated peer reviews \cite{yuYourPaperBeing2025, zhouScholarLensTrackingGrowing2026, duarte2026semdetect}. These approaches often rely on substantial resources, such as reviews regenerated by LLMs from the same paper or likelihood features computed with a 7B-parameter LLM \citep{yuYourPaperBeing2025, duarte2026semdetect}. Rather than competing with these detectors, we aim to provide lightweight linguistic features that complement them at a marginal computational cost, without requiring LLM inference or API access.

In this paper, we propose a novel method for AI-generated text detection that heavily relies on linguistic analysis, and leverages subjectivity encoded at two levels: language \emph{meaning} (semantics) and language \emph{use} (pragmatics). 
At the semantic level, \emph{evaluative language} and \emph{modality} provide access into the speaker's preferences, normative judgments, and epistemic commitments, capturing different dimensions and degrees of subjectivity.\footnote{As discussed in Section \ref{related_gonzalo}, the linguistic literature on evaluative language has traditionally focused on predicates of personal taste \citep[a.m.o.]{umbachEvaluativePredicatesFun2020, stojanovicTalkingTasteDisagreement2007} and, to a lesser extent, on moral predicates \citep{stojanovicAreMoralPredicates2023}. Whether modal expressions should be treated as evaluative in the same sense is more controversial (although, see \citealp{lasersohnSubjectivityPerspectiveTruthTheoretic2016, stephensonJudgeDependenceEpistemic2007, silkDiscourseContextualismFramework2016}).} At the pragmatic level,  speech acts capture  speaker's communicative intentions  (\citealp{austinHowThingsWords1962,searleSpeechActsEssay1969}). 
We hypothesize that these two sources of information provide complementary signals for distinguishing human from AI-generated text. Based on these features, we trained a GRU-based classifier and evaluated their contribution to AI-generated text detection,  both on their own and in combination with an existing detector.

On its own, our full model, depicted in Figure~\ref{pipeline}, achieves an AUROC of 0.9566 and a TPR of 46.8\% at 1\% FPR on the AI Peer Review Detection Benchmark (APRDB;~\citealp{yuYourPaperBeing2025}). Combining its output with the state-of-the-art detector \textit{Sem-Detect}~\citep{duarte2026semdetect} through a simple late fusion raises the TPR at 1\% FPR of \textit{Sem-Detect} on APRDB from 0.874 to 0.985.

The main findings show that speech-act information provides a strong signal for distinguishing human from AI-generated reviews. This contribution arises both from the distribution of individual speech acts and from their sequential patterns. Subjectivity scores also contribute substantially to the model's predictions, providing complementary information to the speech-act features.

The main contributions of our study are as follows. First, we show that linguistic subjectivity, through its encoding of intentions, preferences, and attitudes (traditionally regarded as distinctly human \citep{portnerModality2009, fodorLanguageThought1975}, provides a useful signal for distinguishing human from AI-generated text. Second, the method goes beyond detection by providing explainable insights into the linguistic differences between human and AI-generated texts. Finally, our approach is lightweight: it relies only on two RoBERTa-base models for annotation and a small GRU classifier, and needs only the review itself, not the reviewed paper. It can therefore be combined with existing detectors at a negligible additional cost to complement and improve their performance, as demonstrated by a simple late fusion with \textit{Sem-Detect}~\citep{duarte2026semdetect}.

\section{Related Works}

\subsection{Related Works on Generated Text Classifiers}

The field of detecting generated peer reviews encompasses three main categories of detectors:

- Active detectors are based on text watermarking~\cite{giboulot2024watermax} which can be extremely accurate since the generated text is modified during or after the generation to embed a watermark designed to be easily detectable. Nonetheless, these solutions are associated with practical constraints since they are inefficient if the watermark is not embedded beforehand, or if an adversary tries to remove it~\cite{tarhini:hal-05272053}.

- On the other hand, passive detectors can rely on metrics quantifying the text predictability which is larger when the text is generated. Popular metrics are the perplexity or the cross-perplexity in Binoculars~\cite{hans2024spotting}, the log-likelihood, the token rank~\cite{gehrmann2019gltr}. More elaborated detectors such as {\it detect-GPT}~\cite{mitchell2023detectgpt} or {\it Fast-detect-GPT}~\cite{bao2024fast} uses informational outputs from perturbations of the text to improve the detection. The rational here is that generated texts, once perturbed, will be more affected than human texts which are known to be more erratic and consequently less sensitive to perturbations. %\patrick{cite binoculars} ~\cite{} 

- The last category encompasses passive detectors which use side-information. For generated peer review detection, this side-information is provided by the paper associated to the review. {\it Ancho}r~\cite{yuYourPaperBeing2025} directly compare the test review with a review directly generated from the paper. The comparison is performed by computing the $\cos$ similarity between an embedding of the generated review (the anchor), and the embedding of the tested review. The two embeddings are closer if the tested review is generated than if written by a human. 

A noteworthy similar and competitive approach, but focusing on segments of the review, is proposed by {\it Sem-Detect}~\cite{duarte2026semdetect}. It provides 9 features. The text segments are first categorized into different types and the ones related to ``evaluation'', ``constructive input'' and ``clarification'' are kept. Then five semantic features capturing the closest similarity between embeddings of segments of the test review and segments of regenerated reviews belonging to the same category, are then computed. Four other textures generic features represented by perplexity, entropy, likelihood and {\it Fast-detect-GPT} score, are also computed.

\subsection{Related Works in Linguistics}

\begin{figure*}[th]
    \centering
    \includegraphics[width=\textwidth]{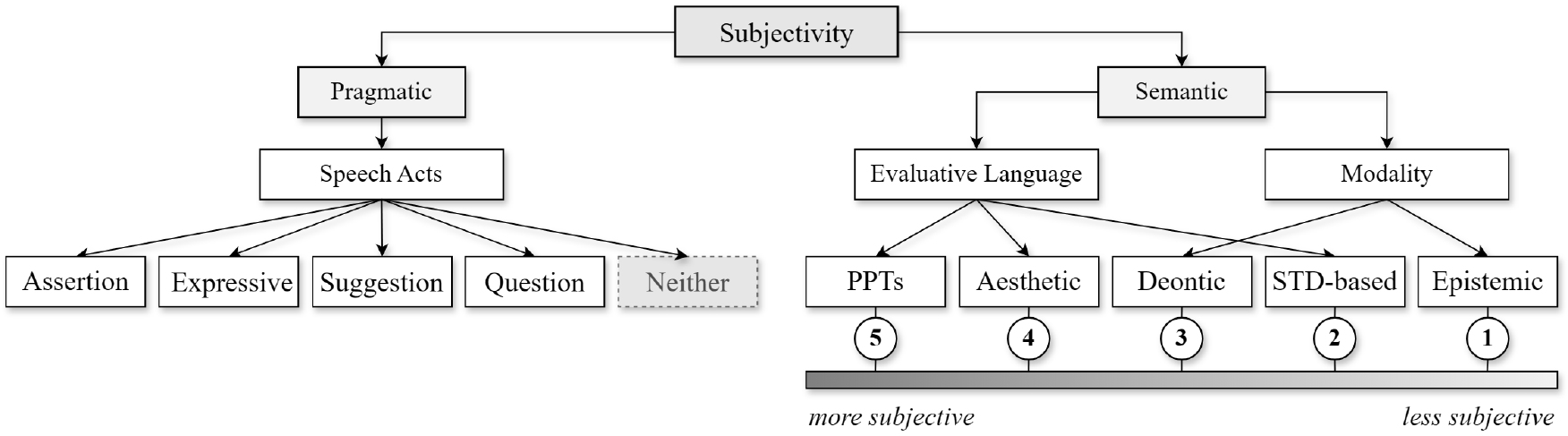}
    \caption{The {\it SAGE} linguistic framework for subjectivity. The pragmatic dimension concerns speech acts, while the semantic dimension covers evaluative language and modality, organized along a five-degree subjectivity scale.}
    \label{schema}
\end{figure*}

Subjectivity broadly concerns the expression of a speaker's perspective, attitudes, and evaluations. It can be expressed through linguistic choices at the lexical level (\textit{tasty, beautiful}; \citealp{lasersohnContextDependenceDisagreement2005, mcnallyAestheticAdjectives2017}) and through modal expressions (\textit{possible, necessary}; \citealp{portnerModality2009, nuytsOxfordHandbookModality2018}). Subjectivity can also be conveyed through speech acts, such as assertions, questions, and exclamatives, which express communicative intentions (\citealp{searleSpeechActsEssay1969, portner2018}).

Speech act annotation has been used in a variety of NLP and discourse-analysis tasks, including dialogue act classification \citep{stolckeDialogueActModeling2000, buntISOStandardDialogue2010}, conversational analysis \citep{asherDiscourseStructureDialogue2016}, argumentation mining \citep{budzynskakatarzynaArgumentMiningDialogue2014}, and the analysis of social media messages for crisis management \citep{laurentiGiveMeYour2022}. %Such annotations provide information about the communicative function of an utterance. 

More recently, \citet{maitraDialogueActsLens2025} used dialogue act annotation to compare LLM-generated and human conversational turns, showing significant differences in their distributions. Their study, however, focuses on telephone conversations from the Switchboard DialogAct corpus,\footnote{\url{https://huggingface.co/datasets/cgpotts/swda}} rather than scientific peer reviews. They report that LLMs produce more opinion statements and questions than humans, whereas in our data, AI-generated reviews consist almost exclusively of assertions, and human-written reviews receive higher subjectivity scores (see Section~\ref{corpus_char}). To the best of our knowledge, speech act annotation has not yet been systematically investigated as a feature for detecting AI-generated text. The closest approach is \citet{duarte2026semdetect}, discussed in Section 2.1. 

On the other hand, although not extensively used, scores of subjectivity have recently been employed as features in text classification.

\citet{mindnerClassificationHumanAIGenerated2023} were the first to use subjectivity as a feature for distinguishing human from AI-generated texts. They use TextBlob's lexicon-based subjectivity score, which ranges from 0 (objective) to 1 (subjective) and is derived from human annotations.\footnote{\url{https://textblob.readthedocs.io/en/dev/}} At the phrase level, scores are computed by averaging the scores of the relevant lexical items. \citet{zhouScholarLensTrackingGrowing2026} subsequently included the same TextBlob score among ten linguistic features for detecting AI-generated reviews and meta-reviews. However, as its lexicon is primarily adjective-oriented, it is limited in its ability to capture subjectivity expressed through other linguistic resources, such as modality. VAGO \citep{icardVAGOOutilLigne2022} relies on a theoretically motivated typology of vague expressions, including modality. However, its proportional measure of subjectivity cannot capture differences in degree between expressions with the same proportion of subjective markers. Although VAGO has been used to characterize human- and AI-generated press-like articles \citep{icardMeasuringVaguenessBasedPersuasion2026}, it has not yet been applied to distinguishing AI-generated from human-written peer reviews.

\section{Subjectivity: semantic and pragmatic dimensions}
\label{related_gonzalo}

Natural languages provide a variety of resources for expressing a speaker's perspective toward some content. As illustrated in Figure~\ref{schema}, we conceptualize subjectivity along two dimensions: a pragmatic dimension, concerning language \emph{use}, and a semantic dimension, concerning linguistic \emph{meaning}. {\it SAGE} captures these dimensions through two complementary sources of information: sentence-level speech act classification and subjectivity scores. The latter are predicted by a transformer trained on statements annotated according to a five-degree scale of subjectivity, as detailed in Section~\ref{ann-pipe}.

\subsection{Pragmatic dimension}

Following \citet{searleSpeechActsEssay1969}, we characterize speech acts in terms of the communicative function or intention of an utterance. For the annotation scheme used in this study, we distinguish four speech act categories: \texttt{Assertions}, \texttt{Questions}, \texttt{Suggestions}, and \texttt{Expressives} (Figure \ref{schema}). These categories capture distinct attitudes toward the propositional content of an utterance: committing to it, as in \texttt{Assertions} \ref{assertion}; requesting information about it, as in \texttt{Questions} \ref{question}; requesting or suggesting that it be realized, as in \texttt{Suggestions} \ref{suggestion}; or expressing internal states, as in \texttt{Expressives} \ref{expressive}.\footnote{On the complexities of speech act type and form-function mismatches, see \citet{ginzburgInteractiveStanceMeaning2012} and \citet{lauer2013towards}. On the effect of speech acts on conversations, see \citet{farkasReactingAssertionsPolar2010, portner2018}, and references therein.} We also include a \texttt{Neither} class for phrases that do not instantiate any of these speech act categories, such as titles, separators \ref{ex8}, or references to sections.

\ex. \label{assertion} This paper introduces a nonlinear dynamical model.\footnote{All examples are drawn from the AI Peer Review Detection Benchmark (APRDB;~\citealp{yuYourPaperBeing2025})}

\ex. \label{question} How does it look visually?

\ex. \label{suggestion} I think it would be good to included these for reference.

\ex. \label{expressive} I am happy with the responses.

\ex. \label{ex8} \#\#\# Strengths

\subsection{Semantic dimension}

Subjectivity at the semantic level can be encoded through perspectival expressions, including \emph{evaluative language} and \emph{modality} (Figure \ref{schema}).

Evaluative language encompasses a variety of expressions that can be graded from more to less subjective. At the more subjective end of this scale are {\it Predicates of Personal Tastes} (PPTs), such as \ref{tst1}. Paradigmatic examples in the semantic literature include \textit{tasty} and \textit{fun}, which can be embedded under \textit{find}, as in \textit{I find this tasty} \citep[a.m.o.]{kolbelIIIFaultlessDisagreement2004,lasersohnContextDependenceDisagreement2005,stephensonJudgeDependenceEpistemic2007,saeboJudgmentAscriptions2009}.

\ex. \label{tst1} The paper \underline{is a joy to read}.

Because PPTs are grounded exclusively in the speaker's individual preferences, they have often been distinguished from other types of evaluative predicates, such as aesthetic predicates \citep{hirvonenPredicatesPersonalTaste2014, mcnallyAestheticAdjectives2017}. Unlike PPTs, aesthetic predicates such as \ref{aesthetic} can be evaluated against both individual preferences and shared aesthetic standards or norms. 

\ex. \label{aesthetic} The idea of the paper \underline{is elegant}.

At a lower degree of subjectivity, \emph{standard-based predicates} as \ref{std-based} qualify an entity, event, or state relative to a standard of adequacy, quality, or success. Like aesthetic predicates, they may be evaluated against standards shared across speakers, but unlike aesthetic predicates, these standards do not interact with individual preferences. 
 \ex. \label{std-based} The effect [...] is not \underline{successfully} demonstrated.

While evaluative language concerns the assessment of entities, events, or states, \emph{modality} concerns the qualification of the propositional content itself in terms of possibility, likelihood, necessity, or obligation, expressed through modal auxiliaries and adverbs such as \emph{may}, \emph{must}, \emph{probably}, and \emph{necessarily} (e.g. \textit{it is probably raining outside}).

The linguistic literature on modality is extremely vast, and it would be impossible to discuss it comprehensively here (see, for instance, \citealp{kratzerWhatMustCan1977,kratzerNotionalCategoryModality1981,hacquardAspectsModality2006,portnerModality2009,rubinstein2012roots,nuytsOxfordHandbookModality2018} and references therein). For present purposes, we broadly distinguish between \emph{epistemic} and \emph{deontic} modality. Epistemic modality concerns the possibility, likelihood, or necessity of a state of affairs as evaluated against available evidence, as in \ref{epistemic_ex}. Deontic modality, in turn, concerns what is permitted, required, or expected according to a set of norms or normative standards, as in \ref{deontic_ex}. 

\ex. \label{epistemic_ex} higher values \underline{might} lead to faster convergence.

\ex. \label{deontic_ex} Data \underline{should} also be leveraged. 

We treat deontic modality as more subjective than both standard-based modifiers and epistemic modality because the relevant norms may depend on the speaker's preferences, although to a lesser extent than in the case of PPTs and aesthetic predicates. Epistemic modality, by contrast, concerns assessments that can in principle be settled by factual evidence and is therefore treated as the least subjective form of modality (see \citealp{portnerModality2009} for discussion). Figure~\ref{schema} illustrates these distinctions and their relative position along the subjectivity scale.

In our study, this five-degree scale provides the linguistic framework for estimating subjectivity, but its categories are not directly annotated in the review corpus. Instead, continuous subjectivity values are predicted by a transformer model trained on a separately annotated corpus (Section~\ref{ann-pipe}).

\section{Annotation Pipeline}\label{ann-pipe}

Our approach consists of three successive processing stages, followed by a classification head. The complete processing pipeline is illustrated in Figure~\ref{pipeline}.

\paragraph{Segmentation.} We used \texttt{PySBD}, a rule-based sentence boundary disambiguation tool that identifies sentence boundaries using linguistic rules and contextual patterns.\footnote{\url{https://github.com/nipunsadvilkar/pySBD}}

\paragraph{Annotation of Speech Acts.} To classify the speech acts described in Section~\ref{related_gonzalo}, we manually annotated 400 held-out reviews sampled from the train subset of the AI Peer Review Detection Benchmark \cite{yuYourPaperBeing2025}. The 6,930 resulting sentences were labelled as \texttt{Assertion}, \texttt{Question}, \texttt{Suggestion}, \texttt{Expressive} or \texttt{Neither} by two annotators (Cohen's $\kappa = 0.775$ on 50 double-annotated reviews; 92.2\% raw agreement).

We fine-tuned RoBERTa-base \citep{liuRoBERTaRobustlyOptimized2019a} on this data for five epochs (learning rate $2\times10^{-5}$, batch size 16, weight decay 0.01) and selected the checkpoint with the highest validation macro-F1\footnote{F1 score averaged over the classes.}, which achieved 97.69\% accuracy and 84.04\% macro-F1. This model was then used to annotate the full corpus.

\paragraph{Subjectivity Estimation.} We used the RoBERTa model fine-tuned as a regressor by \citet{freijedoAduna2026degrees} to automatically assign subjectivity scores to sentences in our corpus that were not labeled as \texttt{Neither}. We selected this model because it was trained using an annotation scheme closely related to that described in Section~\ref{related_gonzalo}. Specifically, it was trained on a corpus of Reddit posts in which subjective statements were classified on a five-point scale of subjectivity, with discrete values ranging from 1 (least subjective) to 5 (most subjective).\footnote{The potential effects of the register difference between the training and target data are discussed as a limitation in Section~\ref{limitations}} Because the model was trained as a regressor, it outputs continuous subjectivity scores rather than discrete class labels. In \citet{freijedoAduna2026degrees}, the model reached an MAE of 0.10 on the normalized scale, with Pearson $r = 0.856$ and Spearman $\rho = 0.830$ on held-out data, supporting the use of its continuous outputs as degrees of subjectivity. We used these scores as a feature for the classification task.

\subsection{Corpus Characteristics}\label{corpus_char}

Before presenting the detection results, we first describe some key characteristics of the annotated corpus.
First, regarding the distribution of speech acts across human and AI-generated reviews, both groups are dominated by \texttt{Assertions}, but AI reviews use them even more (84.6\% vs.\ 73.5\%), whereas human reviews contain more \texttt{Questions}, \texttt{Suggestions}, and \texttt{Expressives} (Table~\ref{distri}). 

\begin{table}[h!]
\centering
\small
\begin{tabular}{lrr}
\toprule
 Speech Act & Human & AI \\
\midrule
Assertion  & 73.5 & 84.6 \\
Question   & 10.6 &  3.1 \\
Neither    &  7.2 & 10.6 \\
Suggestion &  6.4 &  1.6 \\
Expressive &  2.3 &  0.2 \\
\bottomrule
\end{tabular}
\caption{Speech act distribution (\% of sentences) in human and AI-generated reviews from APRDB \cite{yuYourPaperBeing2025}. All differences are significant (Bonferroni-corrected, $\alpha = .05$).}
\label{distri}
\end{table}

Secondly, regarding subjectivity scores, we also assign scores to \texttt{Questions}, as rhetorical or \emph{biased} questions can convey implicit asserted content and the speaker's attitude toward the expected answer. For instance, \ref{q1} challenges the usefulness of the sentence by suggesting that it is redundant.

\ex. \label{q1} Isn't this sentence redundant?

Despite these cases, \texttt{Questions} have the lowest median subjectivity score among the speech-act categories (1.662), while \texttt{Expressives} have the highest (2.863). This pattern is broadly consistent with our annotation scheme, in which \texttt{Expressives} are expected to be the most subjective speech acts.

\begin{figure}[h!]
    \centering
    \includegraphics[width=0.40\textwidth]{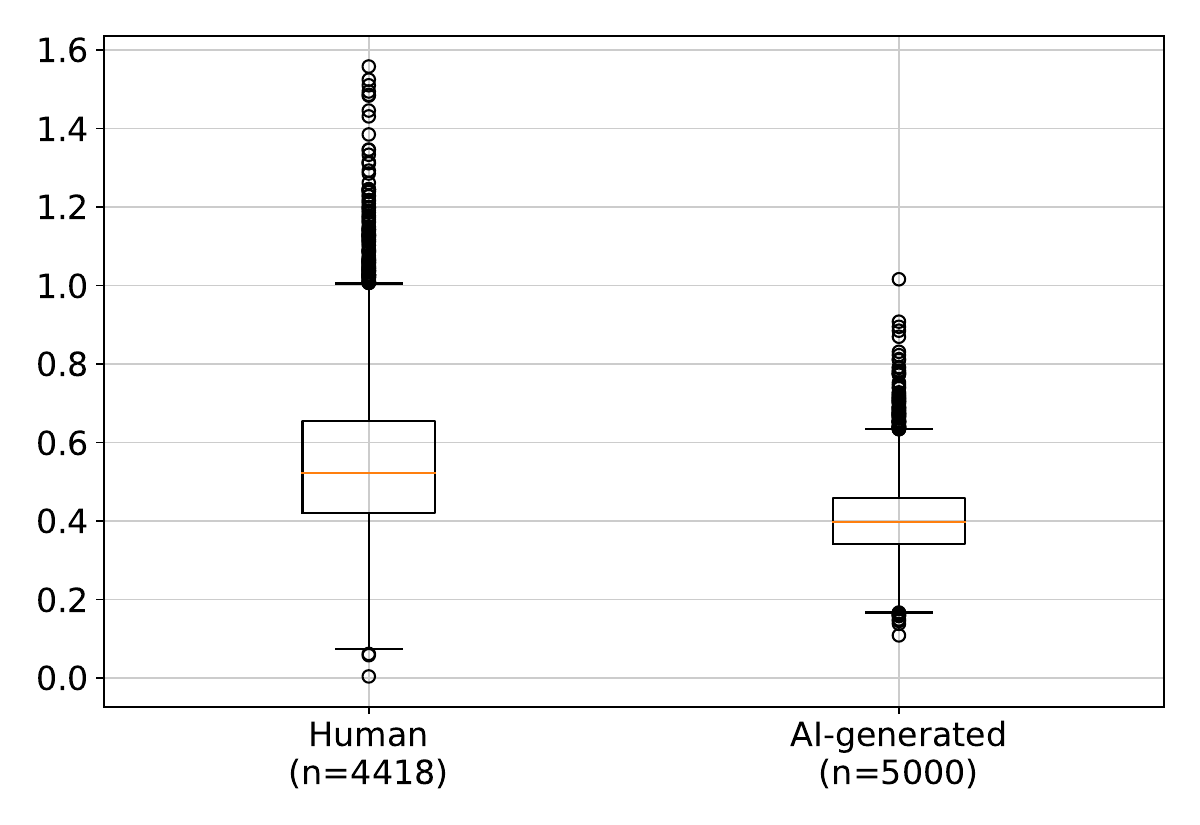}
    \caption{Subjectivity variability in human and AI-generated reviews}
    \label{fig:subj_std_human_ai}
\end{figure}

Finally, to capture within-review variability, we computed the standard deviation of sentence-level subjectivity scores for each review. As shown in Figure~\ref{fig:subj_std_human_ai}, human reviews showed higher variability than AI-generated reviews (median $\mathrm{subj\_std}=0.522$ vs.\ $0.397$; Cliff's $\delta=0.496$, $p<0.001$), indicating greater variation in subjectivity across sentences.

\section{Experiments} \label{experiments}
\subsection{Dataset }
%------------------------------------------------------------------------------

% \patrick{on est pas obligé d'être aussi précis ici, la base est bien décrite dans le papier~\citealp{yuYourPaperBeing2025}}.
\paragraph{APRDB.}
We mainly use the AI Peer Review Detection Benchmark (APRDB;~\citealp{yuYourPaperBeing2025}), a large-scale dataset of human and AI-generated peer reviews released by Intel Labs. It covers reviews from ICLR and NeurIPS and includes texts generated by multiple large language models. The benchmark provides separate train, test, and extended subsets for evaluating AI-generated review detection.

\paragraph{SD-DB}
We also use SD-DB, the dataset associated with the Sem-Detect framework \citep{duarte2026semdetect}, mostly for evaluate the data drift of our model. It comprises three classes of peer reviews from papers sampled from OpenReview: original human-written reviews, fully AI-generated reviews produced by multiple LLMs, and human-written reviews refined by LLMs.

\subsubsection{Dataset Splits}

\paragraph{\texttt{test} split.}
We use the \texttt{test} split of the dataset. It contains 9\,424 examples, with 46.9\,\% human-written reviews and 53.1\,\% AI-generated reviews. The detailed breakdown is shown in Table ~\ref{tab:splits}.

\paragraph{\texttt{train} split.}
For model training, we use the \texttt{train} split (8\,026 examples), consisting exclusively of reviews from 2021, with 37.7\,\% human-written reviews and 62.3\,\% AI-generated reviews. The detailed breakdown is also shown in Table ~\ref{tab:splits}.

\begin{table}[t]
\scriptsize
\centering
\begin{tabular}{lrr}
\toprule\textbf{Variable} & \textbf{\texttt{test}} & \textbf{\texttt{train}} \\
\midrule
{\textit{Generator}} \\
Gemini  & 20.9\% & 19.3\% \\
Llama   & 20.4\% & 19.4\% \\
Qwen    & 20.2\% & 19.6\% \\
Claude  & 19.5\% & 20.8\% \\
GPT-4o  & 19.1\% & 20.9\% \\
\hline
{\textit{Label}} \\
Human (0) & 46.9\% & 37.7\% \\
AI (1) & 53.1\% & 62.3\% \\
\hline
{\textit{Year}} \\
2021 & 22.3\% & 100\% \\
2020 & 8.6\% &  --- \\
2019 & 18.0\% & --- \\
2018 & 17.5\% & --- \\
2017 & 17.3\% & --- \\
2016 & 16.4\% & --- \\
\hline
{\textit{Conference}} \\
NeurIPS & 54.4\% & 49.8\% \\
ICLR & 45.6\% & 50.2\% \\
\bottomrule
\end{tabular}
\caption{Distribution of the \texttt{test} ($n=9.424$) and \texttt{train} ($n=8.026$) splits.}
% , drawn from the database Ai-Peer-Review-Dataset-Benchmark ~\cite{yuYourPaperBeing2025}
\label{tab:splits}
\end{table}

\begin{table*}[ht]
  \centering
  \scriptsize
  \begin{tabular}{lrrrrrrrr}
    \toprule
     & \multicolumn{2}{c}{AUROC} & \multicolumn{2}{c}{TPR@1\%FPR} & \multicolumn{2}{c}{TPR@0.1\%FPR} & \multicolumn{2}{c}{Accuracy} \\
    \cmidrule(lr){2-3} \cmidrule(lr){4-5} \cmidrule(lr){6-7}
    Condition & Value & $\Delta$ & Value & $\Delta$ & Value & $\Delta$ &  Value & $\Delta$ \\
    \midrule
    Permutation of {\bf subjectivity} & 0.8596 & +0.0953 & 0.1762 & +0.2826 & 0.0758 & +0.1326 & 0.7494 & +0.1312 \\
    Permutation of {\bf speech acts} & 0.8355 & +0.1211 & 0.1326 & +0.3365 & 0.0203 & +0.1667 & 0.7527 & +0.1299 \\
    Permutation of {\bf \texttt{assertion}} & 0.8404 & +0.1162 & 0.1399 & +0.3292 & 0.0248 & +0.1622 & 0.7565 & +0.1261 \\
    Permutation of {\bf \texttt{neither}} & 0.9205 & +0.0362 & 0.2474 & +0.2218 & 0.0815 & +0.1054 & 0.8378 & +0.0448 \\
    Permutation of {\bf \texttt{question}} & 0.9291 & +0.0275 & 0.3817 & +0.0874 & 0.1555 & +0.0314 & 0.8448 & +0.0378 \\
    Permutation of {\bf \texttt{suggestion}} & 0.9325 & +0.0241 & 0.3412 & +0.1279 & 0.1156 & +0.0714 & 0.8491 & +0.0335 \\
    Permutation of {\bf \texttt{expressive}} & 0.9511 & +0.0055 & 0.4374 & +0.0317 & 0.1761 & +0.0108 & 0.8751 & +0.0075 \\
    {\bf Shuffled} (train + test) & 0.9079 & +0.0487 & 0.2917 & +0.1774 & 0.0761 & +0.1108 & 0.8191 & +0.0635 \\
    {\bf Full} model & 0.9566 & -- & 0.4691 & -- & 0.1869 & -- & 0.8826 & -- \\
    \bottomrule
  \end{tabular}
  \caption{Feature importance ~\cite{breiman2001random} by permutation with retraining. The category indicator is sampled at its own marginal rate in both train and test, so prevalence is preserved and only the association with the label is destroyed; the model is then retrained (5 seeds). Feature count and model capacity are identical across conditions.
  }
  \label{tab:speech_acts_noise0}
\end{table*}

\subsection{Sequential GRU Model training of {\it SAGE} features}
%------------------------------------------------------------------------------

Our classifier exploits the chronological order of sentences within each review. Instead of aggregating statistics into a fixed vector, we model each review as a sequence of sentences read by a GRU (\emph{Gated Recurrent Unit}, \citealp{cho2014learning}) layer.

In this model, a \emph{block} is a sequence: each sentence is one time step, in chronological order within the review. At each step, the model receives a feature vector describing that sentence. The GRU models sentences in order and maintains a hidden state that summarizes ``what has happened so far.'' Sequence length is therefore the number of sentences in the block, \emph{variable} from one review to another. The shape of a block is $(L_b, 6)$, where $L_b$ is the sequence length (number of sentences). Sentence order in the dataset is preserved and correctly sequential.

At each time step, the feature vector has 6 dimensions:
\begin{itemize}
    \item 
    one-hot encoding of the speech-act categories (5 dimensions) and
    \item 
    one raw subjectivity value (1 dimension). When the subjectivity score is missing (sentence classified as \texttt{Neither}), the value is imputed to a constant.
\end{itemize}

The GRU is trained on \texttt{train} set (see Table~\ref{tab:splits}), using the same annotated reviews. Evaluation uses the same balanced subset of the \texttt{test} split: 8\,848 reviews (4\,424 human-written, 4\,424 AI-generated).

\subsection{Overall classification performance}

On the APRDB test subset, {\it SAGE} reaches an AUROC of 0.964 and an accuracy of 0.901. At a false-positive rate of 5\%, it detects 81.4\% of AI-generated reviews; at 1\%, this proportion falls to 54.6\%. Performance is homogeneous across four of the five generators, with AUROCs between 0.96 and 0.98, while Llama 3.1 70B is a clear outlier, with an AUROC of 0.93 and a TPR@1\% of 0.28. Across conference editions, AUROCs range from 0.938 to 0.99. Detailed results by generator and conference edition, together with ROC curves, are reported in Appendix~\ref{sec:overall_performance}.

\subsection{Ablation study and analysis}

We conducted an ablation analysis through permutation~\cite{altmann2010permutation} to assess the contribution of our two annotation levels to the model's predictions. 

The first level consists of speech act classifications, while the second assigns subjectivity scores to each sentence. As shown in Table~\ref{tab:speech_acts_noise0}, removing speech act information has a greater impact on model performance than removing subjectivity scores, with AUROC decreasing from 0.96 to 0.84 and 0.86, respectively. This indicates that speech acts provide the stronger signal for distinguishing human- from AI-generated reviews. Nevertheless, subjectivity scores also have a substantial impact on performance, suggesting that they provide complementary information to the speech act annotations. We can note that the importance of the different categories  of speech acts follows roughly the proportions of these categories in the corpus (see Table~\ref{distri}). 

Finally, the order of speech acts also contributes to model performance: shuffling their order reduces AUROC from 0.96 to 0.91. This means that the performance of the detector is linked to the possibility to exploit the sequence of categories and subjectivity scores.

\begin{table*}[t]
  \centering
  \scriptsize
  \begin{tabular}{lllccc}
    \toprule
    Detector & Training data & Accuracy & AUROC & TPR@1\% \\
    \midrule
    {\it Binoculars}~\citep{hans2024spotting} & zero-shot & 0.8203 & 0.9836 & 0.7676 \\
    {\it Fast-DetectGPT}~\citep{bao2024fast} & zero-shot  & 0.7900 & 0.7304 & 0.3974 \\
    {\it SAGE} & APRDB (train)  & 0.8826 & 0.9566 & 0.4691 \\
    {\it Sem-Detect}~\citep{duarte2026semdetect} & SD-DB & 0.9424 & 0.9801 & 0.6999 \\
    \bottomrule
  \end{tabular}
  \caption{Comparison of detectors on APRDB ~\cite{yuYourPaperBeing2025}. Detectors are evaluated on a subsets of the APRDB test split.}
  \label{tab:detector_comparison}
\end{table*}

\begin{table*}[h!]
  \centering
  \scriptsize
  \begin{tabular}{lcccccccc}
    \toprule
     & \multicolumn{4}{c}{Test: APRDB} & \multicolumn{4}{c}{Test: SD-DB} \\
    \cmidrule(lr){2-5} \cmidrule(lr){6-9}
    Detector & AUROC & TPR@1\% & TPR@5\% & Acc$_{\max}$ & AUROC & TPR@1\% & TPR@5\% & Acc$_{\max}$ \\
    \midrule
    {\it SAGE}-APRDB & 0.9566 & 0.4691 & 0.7530 & 0.8826 & 0.7208 & 0.2060 & 0.3510 & 0.7277 \\
    {\it SAGE}-SD-DB & 0.7971 & 0.0509 & 0.1661 & 0.7496 & 0.9999 & 0.9970 & 0.9980 & 0.9965 \\
    {\it Sem-Detect} & 0.9801 & 0.6999 & 0.9059 & 0.9424 & 0.9987 & 0.9670 & 0.9990 & 0.9867 \\
    \midrule
   \\
    Hybrid (SVM), trained on APRDB & 0.9984 & 0.9845 & 0.9956 & 0.9875 & 0.9996 & 0.9850 & 1.0000 & 0.9902 \\
    Hybrid (SVM), trained on SD-DB & 0.9982 & 0.9756 & 0.9911 & 0.9842 & 0.9998 & 0.9960 & 1.0000 & 0.9951 \\
    \bottomrule
  \end{tabular}
  \caption{
  Cross-dataset evaluation. Each detector is trained on one dataset and evaluated on both. {\it SAGE}-APRDB is trained on APRDB; {\it SAGE}-SD-DB and Sem-Detect~\citep{duarte2026semdetect} are trained on SD-DB. }
  \label{tab:cross_dataset}
\end{table*}

\subsection{Comparison with SOTA detectors}

Table~\ref{tab:detector_comparison} compares {\it SAGE} with two zero-shot detectors and a domain-specific supervised detector. {\it SAGE} achieves an accuracy of 0.901, below {\it Sem-Detect} (0.942), and an AUROC of 0.964, close to Binoculars (0.984) and {\it Sem-Detect} (0.980), and well above {\it Fast-DetectGPT} (0.730).

Unlike these methods, {\it SAGE} only uses the sequence of speech-act categories and subjectivity scores of the target review. Binoculars requires token-level likelihoods from two 7B-parameter LLMs, while {\it Sem-Detect} uses AI-generated reference reviews of the same paper, LLM-based claim extraction, and likelihood features from a 7B-parameter LLM. Thus, review-level subjectivity features provide a competitive detection signal while remaining explainable.

\paragraph{Cross-dataset generalization.}
Table~\ref{tab:cross_dataset} evaluates each detector on APRDB and SD-DB \cite{duarte2026semdetect}. In distribution, {\it SAGE} is on par with {\it Sem-Detect}: trained on SD-DB, it reaches an AUROC of 0.9999 and a TPR@1\% of 0.997, against 0.9987 and 0.967 for {\it Sem-Detect}.%, although TPR@1\% on this test set rests on about four false positives. 

Across benchmarks, both {\it SAGE} models degrade sharply. {\it SAGE}-APRDB drops to an AUROC of 0.721 on {\it Sem-Detect}, and {\it SAGE}-SD-DB to 0.797 on APRDB. The transfer is asymmetric at low false-positive rates: {\it SAGE}-APRDB still detects a fifth of AI reviews at 1\% FPR (0.206), whereas {\it SAGE}-SD-DB detects almost none (0.051).
This sensitivity to the source benchmark is not specific to {\it SAGE}, {\it Binoculars} reaches an AUROC of 0.984 on APRDB (Table~\ref{tab:detector_comparison}).

Since {\it SAGE} models the sequence of speech acts, such a fixed template plausibly provides a structural cue that is learnable in distribution but does not transfer. This is consistent with the near-perfect in-distribution score of {\it SAGE}-SD-DB and its weak transfer to APRDB. 
\emph{Sem-Detect} transfers markedly better, with an AUROC of 0.980 on APRDB.

\subsection{Ensembling with Sem-Detect}

{\it SAGE} and \emph{Sem-Detect} rely on different signals: {\it SAGE} models the sequence of speech acts and subjectivity of a review, while Sem-Detect compares its semantic claims with reference AI reviews. To test whether these signals are complementary, we combine the two detectors in the simplest possible way, without modifying either of them. For each review, we take the output logits of {\it SAGE}-APRDB and of \emph{Sem-Detect} and train a linear SVM on them. Table~\ref{tab:cross_dataset} shows that the hybrid improves over \emph{Sem-Detect} on APRDB, raising TPR@1\%FPR from 0.700 to 0.985 and the best accuracy from 0.942 to 0.988. The gain persists when the SVM is fitted on SD-DB and applied to APRDB without refitting (0.976). To separate the contribution of {\it SAGE} from that of the SVM itself, we apply the same SVM to the \emph{Sem-Detect} prediction alone: its TPR@1\%FPR on APRDB reaches only 0.874. Adding {\it SAGE} therefore brings a further gain of 0.11, which shows that the subjectivity and speech-act features capture information that the semantic comparison of \emph{Sem-Detect} misses. On SD-DB, both detectors are already close to the ceiling, and the hybrid matches \emph{Sem-Detect} within the resolution of the test set.

\section{Conclusions}

We presented a novel approach to detecting AI-generated peer reviews by exploiting subjectivity, an inherently human feature, through linguistic annotations at two levels: speech acts and degrees of subjectivity. Our results show that {\it SAGE} achieves competitive detection performance, ranking second to Sem-Detect despite relying on a substantially simpler approach. On their own, these features already carry a substantial detection signal, despite relying only on two RoBERTa-base models and a small GRU, without any LLM inference or access to the reviewed paper. More importantly, this lightweight design allows them to be combined with existing detection methods at a negligible additional cost to enhance their performance: a simple late fusion of \textit{SAGE} with \textit{Sem-Detect} yields improved detection performance over either method alone.

\section{Limitations}\label{limitations}

\paragraph{Subjectivity estimation.} The model used to assign subjectivity scores was trained on Reddit conversations, whose informal register differs from that of peer reviews, which may limit its generalizability to this domain. As a result, the subjectivity scores may be less reliable for review sentences, and \emph{SAGE}'s performance may be lower than it would be with a model fine-tuned on peer-review data.

\paragraph{Language.}
Our study is restricted to English, the working language of the venues covered by APRDB, and the upstream annotators were applied to English text only. Extending SAGE to other languages would require speech-act and subjectivity annotators for those languages, and the use of speech acts in reviews may itself vary across languages and scientific cultures. Within English, a more pressing issue is that many reviewers write in English as a second language, and AI-text detectors have been shown to flag non-native English writing disproportionately~\citep{liang2023gpt}. We did not evaluate whether the pragmatic profile used by SAGE differs between native and non-native reviewers, which would be necessary before any use that could lead to accusations against reviewers.

\paragraph{Cross-dataset transfer.} \textit{SAGE} is a supervised detector: its decision boundary is learned from the distribution it is trained on. Applied without adaptation to a benchmark built with different generators and prompting strategies, its performance decreases. Trained on APRDB, its AUROC goes from 0.959 in distribution to 0.721 on the SD-DB. %benchmark~\citep{duarte2026semdetect}. 
Conversely, trained on Sem-Detect, it reaches 0.797 on APRDB (Table~\ref{tab:cross_dataset}). Since {\it SAGE} performs strongly on each benchmark when trained in-domain, this gap suggests that the pragmatic features carry signal on both benchmarks. The learned decision boundary, however, remains specific to each generation setup, which differs in generators, prompts and rating conditioning.

\bibliography{custom}

\appendix

\section{Overall classification performance with {\it SAGE} features}
\label{sec:overall_performance}

On the APRDB test subset, {\it SAGE} reaches an AUROC of 0.964 and an accuracy of 0.901 (Table~\ref{tab:SAGE_generator}). At a false-positive rate of 5\%, it detects 81.4\% of AI-generated reviews; at 1\%, this proportion falls to 54.6\%. {\it SAGE} therefore ranks most reviews correctly, but its reliability decreases in the low false-positive regime that matters most in deployment.

\paragraph{Per-generator performance.}
Performance is homogeneous across four of the five generators, with AUROCs between 0.960 and 0.979 and TPR@1\% between 0.552 and 0.684 (Table~\ref{tab:SAGE_generator}). Qwen 2.5 72B is the most detectable generator on every metric. Llama 3.1 70B is a clear outlier: its AUROC drops to 0.931 and its TPR@1\% to 0.284, less than half the average TPR@1\% of the four other generators (0.613). Llama reviews therefore account for much of the performance gap at low false-positive rates.

This pattern is the reverse of the one reported for likelihood-based detectors. \citet{yuYourPaperBeing2025} find Llama reviews to be among the easiest to detect for token-level methods, with Binoculars exceeding 99\% TPR at 1\% FPR. In contrast, their Anchor detector, which relies on semantic similarity to reference reviews, detects only 36.8\% of them at the same operating point. Conversely, GPT-4o, which they find hardest to detect for Binoculars, is not the hardest generator for {\it SAGE}. The error profile of {\it SAGE} across generators resembles that of a content-level detector rather than that of token-level detectors. This suggests that its signal may be complementary to likelihood-based methods.
\begin{figure*}[t]
  \centering
  \includegraphics[width=0.9\textwidth]{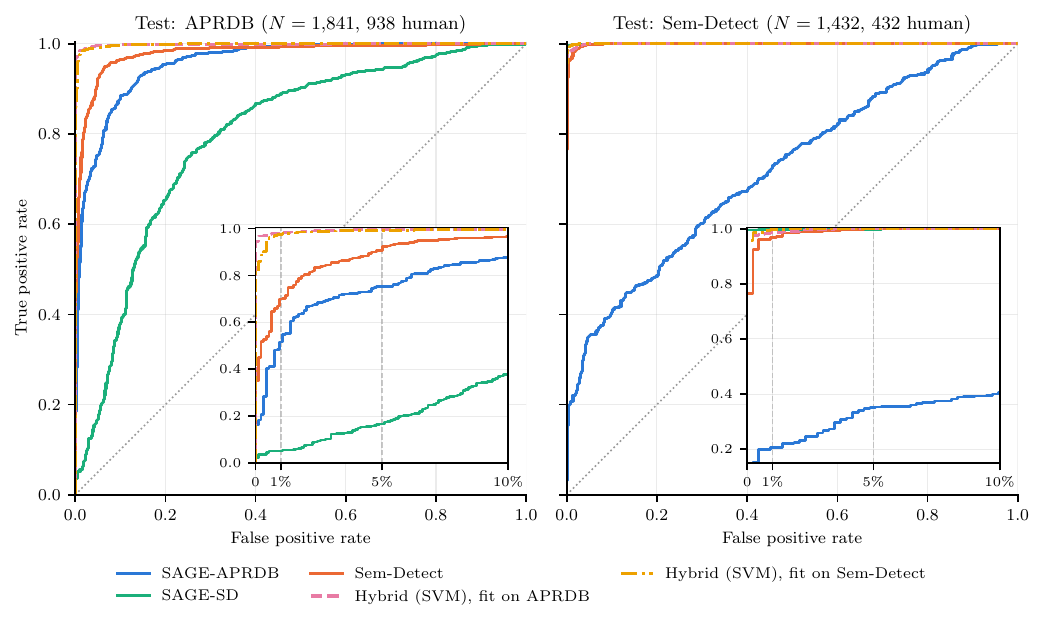}
  \caption{ROC curves of SAGE, Sem-Detect and their SVM hybrid on the APRDB test subset (left) and the Sem-Detect test set (right). SAGE-APRDB and SAGE-SD denote SAGE trained on APRDB and on Sem-Detect, respectively. The hybrid is an SVM fitted on the output logits of both detectors, either on APRDB or on Sem-Detect. Insets zoom on the low false-positive region (FPR $\leq 10\%$); the dotted diagonal corresponds to a random classifier.}
  \label{fig:roc_comparison}
\end{figure*}

\begin{table}[t]
  \centering
  \scriptsize
  \setlength{\tabcolsep}{3pt}
  \begin{tabular}{lcccc}
    \toprule
    Generator & AUROC & TPR@1\% & TPR@5\% & Accuracy \\
    \midrule
    GPT-4o & 0.9766 & 0.5770 & 0.8733 & 0.9264 \\
    Claude Sonnet 3.5 & 0.9755 & 0.6393 & 0.8849 & 0.9210 \\
    Gemini 1.5 Pro & 0.9604 & 0.5515 & 0.8206 & 0.8970 \\
    Qwen 2.5 72B & 0.9788 & 0.6844 & 0.8955 & 0.9270 \\
    Llama 3.1 70B & 0.9309 & 0.2842 & 0.6030 & 0.8589 \\
    \midrule
    All & 0.9566 & 0.4691 & 0.7530 & 0.8826 \\
    \bottomrule
  \end{tabular}
  \caption{Detection performance of \it{SAGE} per generator on the APRDB test subset.}
  \label{tab:SAGE_generator}
\end{table}

\paragraph{Per-edition performance.}
Performance is stable across venues, with pooled AUROCs of 0.966 for ICLR and 0.963 for NeurIPS, and across editions, with AUROCs ranging from 0.938 to 0.990 (Table~\ref{tab:SAGE_edition}). This stability is notable because {\it SAGE} is trained on the APRDB train split, which only covers the 2021 and 2022 editions of both venues~\citep{yuYourPaperBeing2025}. The 2016--2020 editions therefore constitute a temporal shift. This shift also involves changes in review templates and, for ICLR 2017--2018 and NeurIPS 2016--2019, a different data source (ASAP-Review). The 2021 editions, the only ones in our test subset that are also covered by the training split, yield the highest AUROC within each venue (0.990 for ICLR, 0.978 for NeurIPS). Earlier editions show a moderate degradation, with ICLR 2020 as the weakest (AUROC 0.938). ICLR 2020 is also the only ICLR edition in our subset whose template adds structured self-assessment fields to the free-text review (reviewer experience, checking of derivations and experiments, thoroughness of reading). These fields may alter the speech-act profile of human reviews.

Overall, {\it SAGE}, trained on two recent editions, maintains an AUROC above 0.93 on reviews written up to five years earlier. This suggests that its discriminative signal can remain useful across editions despite temporal shifts and changes in review templates.

\begin{table}[h!]
  \centering
  \scriptsize
  \setlength{\tabcolsep}{3pt}
  \begin{tabular}{llcccc}
    \toprule
    Conf. & Year & AUROC & TPR@1\% & TPR@5\% & Accuracy \\
    \midrule
    ICLR & 2017 & 0.9563 & 0.3426 & 0.8194 & 0.8963 \\
         & 2018 & 0.9652 & 0.4976 & 0.8180 & 0.9064 \\
         & 2019 & 0.9709 & 0.5316 & 0.8584 & 0.9217 \\
         & 2020 & 0.9377 & 0.3762 & 0.7336 & 0.8552 \\
         & 2021 & 0.9901 & 0.8611 & 0.9470 & 0.9513 \\
    \cmidrule(lr){2-6}
         & \textit{All} & 0.9661 & 0.5452 & 0.8336 & 0.9033 \\
    \midrule
    NeurIPS & 2016 & 0.9440 & 0.4672 & 0.7282 & 0.8629 \\
           & 2017 & 0.9692 & 0.6027 & 0.8539 & 0.9114 \\
           & 2018 & 0.9693 & 0.6732 & 0.8182 & 0.9078 \\
           & 2019 & 0.9695 & 0.5335 & 0.8568 & 0.9066 \\
           & 2021 & 0.9776 & 0.5540 & 0.8779 & 0.9206 \\
    \cmidrule(lr){2-6}
           & \textit{All} & 0.9626 & 0.5290 & 0.7906 & 0.8973 \\
    \bottomrule
  \end{tabular}
  \caption{Detection performance of {\it SAGE} per conference edition on the APRDB test subset.}
  \label{tab:SAGE_edition}
\end{table}

\label{sec:appendix}

\end{document}